\documentclass{article}

\usepackage{arxiv}
\usepackage{url}
\usepackage{fancyhdr}
\usepackage{natbib}
\usepackage{multirow}
\usepackage{diagbox}
\usepackage[T1]{fontenc}
\usepackage{wrapfig}
\RequirePackage{algorithm}
\usepackage[backref]{hyperref}
\usepackage{booktabs}
\usepackage{multirow}
\usepackage{xcolor}
\usepackage{bibentry}


\usepackage{xcolor}
\definecolor{lightgray}{gray}{0.893}
\usepackage{colortbl}

\definecolor{GainRed}{RGB}{180,45,35}
\newcommand{\gain}[1]{\textcolor{GainRed}{\textbf{$\uparrow$\,#1}}}
\newcommand{\best}[1]{\textbf{#1}}

\usepackage{xcolor}         
\usepackage{bm}
\usepackage{amsmath}
\usepackage{amsthm} 
\usepackage{multicol} 

\usepackage{subcaption}

\usepackage{soul}
\usepackage{booktabs}

\usepackage{amssymb}
\usepackage{bm}
\usepackage{amsmath}
\usepackage{amsthm}

\usepackage{times}
\usepackage{helvet}
\usepackage{courier}
\usepackage{xcolor}
\usepackage{booktabs}
\usepackage{multirow}

\usepackage{algorithm}
\usepackage{algpseudocode}
\usepackage{graphicx} 

\title{Autonomous Aerial Manipulation via Contextual Contrastive Meta Reinforcement Learning}

\author{
Lixuan Jin$^{1,2}$, Bingxuan Lan$^{1,2}$,
Xinyi Bao$^{1,2}$, Xiangyuan Xie$^{3}$, Chunjie Zhang$^{3}$, Zheng Chen$^{1,2}$, 
\\ \textbf{Tianshuo Liu$^{1,2}$, Ruijie Tian$^{1,2}$, Jinyu Ru$^{3}$, Gang Wang$^{4}$, Lei Yuan$^{1,2}$, Yang Yu$^{1,2}$}\\   
$^1$ National Key Laboratory of Novel Software Technology, Nanjing University, Nanjing, China\\
$^2$ School of Artificial Intelligence, Nanjing University, Nanjing, China\\
$^3$ Faculty of Robot Science and Engineering, Northeastern University, China\\
$^4$ National Key Lab of Autonomous Intelligent Unmanned Systems, Beijing Institute of Technology, China\\
\texttt{\{jinlx,lanbx,baoxy\}@smail.nju.edu.cn}, \\ \texttt{\{xiexiangyuan,zhangcj,rujingyu\}@mails.neu.edu.cn},\\ \texttt{\{chenz,liuts,tianrj,yuanl,yuy\}@lamda.nju.edu.cn},
\texttt{\{gangwang\}@bit.edu.cn}
}

\begin{document}
\date{}
\maketitle
\begin{abstract}
    Unmanned aerial vehicles (UAVs) are increasingly being deployed in logistics, service robotics, and other real-world applications, creating a growing demand for autonomous payload acquisition and delivery. Existing approaches typically assume pre-attached payloads or rely on specialized grippers, leaving versatile end-to-end aerial delivery largely unresolved, where different payloads induce highly variable flight dynamics, requiring a single policy to adapt online without manual calibration or explicit system identification. To this end, we study \textbf{A}utonomous \textbf{A}erial Manipulation via \textbf{Co}ntextual \textbf{Co}ntrastive Meta Reinforcement Learning (\textbf{\textit{Aco2}}), a fully autonomous aerial delivery setting in which a quadrotor equipped with a lightweight hook continuously picks up, transports, and delivers diverse handle-equipped objects between randomized locations, all without human intervention. First, we design a contextual observation encoder that infers a compact latent context from recent interaction history, enabling the policy to adapt online to payload-dependent dynamics. To further improve the quality of this context, we introduce a contrastive objective that structures the context embedding around task-relevant variations, improving generalization across diverse payloads without requiring explicit system identification. Trained entirely in simulation with extensive domain randomization, \textit{Aco2} can be directly deployed on a physical quadrotor without real-world fine-tuning. Experiments show that it enables unattended delivery of handle-equipped objects that are relatively heavy or bulky for the vehicle, demonstrating the practical potential of \textit{Aco2} for scalable aerial logistics, emergency supply delivery, and industrial material transport by leveraging the fast and obstacle-free mobility of UAVs. An overview video is available at \url{https://youtu.be/DLuZz9xxrJQ}.

\end{abstract}

\section{Introduction}
Benefiting from high maneuverability and largely obstacle-free airspace, unmanned aerial vehicles (UAVs) have been increasingly adopted across a wide spectrum of civilian and industrial domains, including environmental monitoring~\citep{asher2021unpiloted,chen2026wind}, precision agriculture~\citep{ahmed2026uav,xing2025integrating,manu2026uav,arsenoaia2026sensing}, and infrastructure inspection~\citep{mei2025unmanned,liu2026drone}. Beyond perception and navigation, there is growing interest in UAVs that physically interact with objects and environments~\citep{barakou2024review,deshmukh2025global}, with autonomous object transport emerging as a particularly pressing need in demanding scenarios such as emergency supply delivery~\citep{nisingizwe2022effect} and industrial material handling~\citep{trujillo2019novel}. The payloads encountered in these settings are often heavy, bulky, and irregularly shaped, with widely varying mass and inertial properties that substantially alter the quadrotor's flight dynamics upon attachment\citep{sreenath2013geometric}, demanding a control policy with strong generalization across diverse objects and rapid online adaptation to each new payload.

	Existing work has made considerable progress but typically addresses only part of this problem. A large body of aerial transport research assumes the payload is manually attached before takeoff~\citep{zeng2025decentralized,lorentz2025crazymarl,cao2026flare,wang2025learning}, which not only requires human intervention but also results in a relatively task-specific setting where the controller is rarely exposed to unfamiliar dynamics. Methods that incorporate autonomous grasping or hooking~\citep{antal2025autonomous,antal2025hook,ye2026flyaware,d2026hitch,chen2025aerial} remove this dependency, but typically require modeling, parameter tuning, or controller design, limiting their scalability to diverse objects and deployment conditions. Learning-based approaches have begun to relax these assumptions, yet demonstrations remain limited to simple, lightweight objects such as cups, plush toys, or small parcels~\citep{tucker2026pi,huang2025swooper}, far from the payload diversity that real-world deployment demands.

To address these challenges, we adopt a lightweight passive hook as the end-effector, capable of acquiring any handle-equipped object regardless of its geometry and mass.
Building on this platform, we propose \textbf{A}utonomous \textbf{A}erial Manipulation via \textbf{Co}ntextual \textbf{Co}ntrastive Meta Reinforcement Learning (\textbf{\textit{Aco2}}) to handle the highly variable flight dynamics induced by diverse payloads. Our framework employs a history-conditioned policy that maintains a compact latent embedding for online dynamics inference, with a contrastive auxiliary objective that further structures this embedding by separating representations of different payloads, mitigating the representation collapse common in multi-task learning.
Training the full hooking, transport, and detachment sequence end-to-end presents significant optimization difficulties. We address this through a curriculum learning strategy~\citep{bengio2009curriculum} that decomposes the task into manageable stages. A threshold-based reward formulation balances aggressive task completion against conservative control, and extensive domain randomization across physical and perceptual parameters equips the policy with robustness to the sim-to-real gap. Together, these strategies make the learning of \textit{Aco2} tractable and enable its successful zero-shot transfer to a physical quadrotor.

We comprehensively evaluate the proposed \textit{Aco2} framework across simulated environments and real-world deployments, with the two settings playing
  complementary roles. Leveraging the controlled, quantitative evaluation afforded by simulation, we isolate the effect of the proposed contrastive context learning in producing robust contextual representations
  across diverse payloads. Since real-world experiments are required to verify sim-to-real transfer of the policy under payload variations and hardware constraints, we use them to demonstrate \textit{Aco2}'s capability to perform fully autonomous deliveries by executing the complete pipeline of payload engagement, transportation, and automated detachment. Finally, we conduct ablation studies to verify that our proposed regularization terms and threshold-based rewards are strictly necessary for maintaining physical flight stability and robust policy optimization.

\section{Related Work}
\label{sec:related_work}

\subsection{Aerial Manipulation} 
Aerial manipulation~\citep{khamseh2018aerial,ruggiero2018aerial} and transport aim to enable autonomous delivery across payloads with diverse geometries, masses, and inertial properties.
Several existing suspended load approaches~\citep{sreenath2013trajectory} focus on overcoming in-flight external disturbances by dynamics modeling. They rely on pre-suspended payloads before flight, which cannot address the problem of autonomous attachment. Cooperative cable transport~\citep{sreenath2013dynamics,lee2017geometric,tognon2018aerial} extends this setting to multiple UAVs, yet it remains dependent on pre-hooked setups, unable to address complex docking situations and resulting in relatively task-specific applications.
Another class of methods employs sophisticated controller design~\citep{pounds2011grasping,lippiello2012cartesian, antal2025hook,d2026hitch} and modeling~\citep{mellinger2011design,orsag2013modeling,kim2013aerial,ye2026flyaware} to enable autonomous payload acquisition and manipulation, eliminating the need for manual payload pre-attachment. However, their reliance on explicit models or task-specific controllers makes generalization across diverse payloads nontrivial.
Recent studies have opted for learning-based methods, such as reinforcement learning~\citep{huang2025swooper} and vision-language-action models~\citep{tucker2026pi}, which effectively relax these constraints. However, the manipulated objects in current demonstrations remain relatively simple, leaving open the challenge of robust delivery across diverse payloads.

\subsection{Meta-RL}
 Traditional RL typically optimizes a policy within a fixed environmental configuration, making it prone to overfitting to specific task dynamics. Meta-RL addresses this limitation by
enabling rapid adaptation to unseen tasks through meta-learning how policies solve problems  ~\citep{beck2025tutorial}. 
This makes meta-RL particularly relevant to robotics, where agents must quickly adjust to varying tasks, dynamics, and environmental conditions during deployment.

MAML~\citep{finn2017model}-based methods typically learn an initialization from training tasks and adapt it to a new task through several policy-gradient updates ~\citep{kim2025disentangled,arndt2020meta,yu2020learning,ghadirzadeh2021bayesian}. Such methods require additional model updates in the target environment and typically rely on at least one trajectory for adaptation, which can be costly and risky for safety-critical real-world robotic systems, especially aerial manipulation.
Another category of methods conditions the policy on a context vector inferred from historical data to represent the current task ~\citep{akkaya2019solving,kumar2021rma,he2023learning}. Among these, ~\citep{fu2021towards,wang2023meta} use contrastive losses to improve context quality by learning more discriminative task representations. Yet, to the best of our knowledge, existing contrastive meta-RL methods are predominantly evaluated in simulated benchmarks (e.g., MuJoCo~\citep{todorov2012mujoco}, Meta-World~\citep{yu2020meta}), while real-world meta-RL demonstrations have mainly focused on ground robots, legged locomotion, or dexterous manipulation platforms. 
In contrast, our work demonstrates contrastive meta-RL on a physical quadrotor system for zero-shot aerial manipulation, realizing autonomous delivery across diverse payloads.
\section{Preliminaries}
\label{sec:preliminaries}
We model the aerial manipulation problem as a partially observable Markov decision process (POMDP), defined by the tuple
$\mathcal{M}=(\mathcal{S}, \mathcal{A}, \mathcal{O}, P, R, \Omega, \gamma,\rho_0)$.
Here, $\mathcal{S}$ is the full state space, $\mathcal{A}$ is the action space, $\mathcal{O}$ is the observation space, $P(s_{t+1}\mid s_t,a_t)$ is the transition function, $R(s_t,a_t)$ is the reward function, $\Omega(o_t\mid s_t)$ is the observation function, $\gamma\in[0,1)$ is the discount factor, and $\rho_0 $ is the initial distribution over states.
At each timestep, the agent receives an observation $o_t\in\mathcal{O}$, selects an action $a_t\in\mathcal{A}$ according to its policy. 

We consider a distribution of POMDPs $\{\mathcal{M}_\mu\}_{\mu\sim p(\mu)}$, where each task is defined as
\(
\mathcal{M}_{\mu}
=
(\mathcal{S}, \mathcal{A}, \mathcal{O}, P_{\mu}, R, \Omega, \gamma, \rho_0).
\)
The tasks share the same observation space, observation function, and reward function, while the transition dynamics $P_\mu$ vary with the latent context $\mu$. In our setting, $\mu$ captures unobserved payload and system properties, including object mass, geometry, center-of-mass offset, aerodynamic drag, motor delay, and action delay, among others. Observations are additionally corrupted by noise under a shared observation model. Initial and target locations are randomized across episodes and represented through relative-position observations.
\section{Method}
\label{sec:method}

\begin{figure}[t!] 
  \centering
  \includegraphics[width=0.9\textwidth]{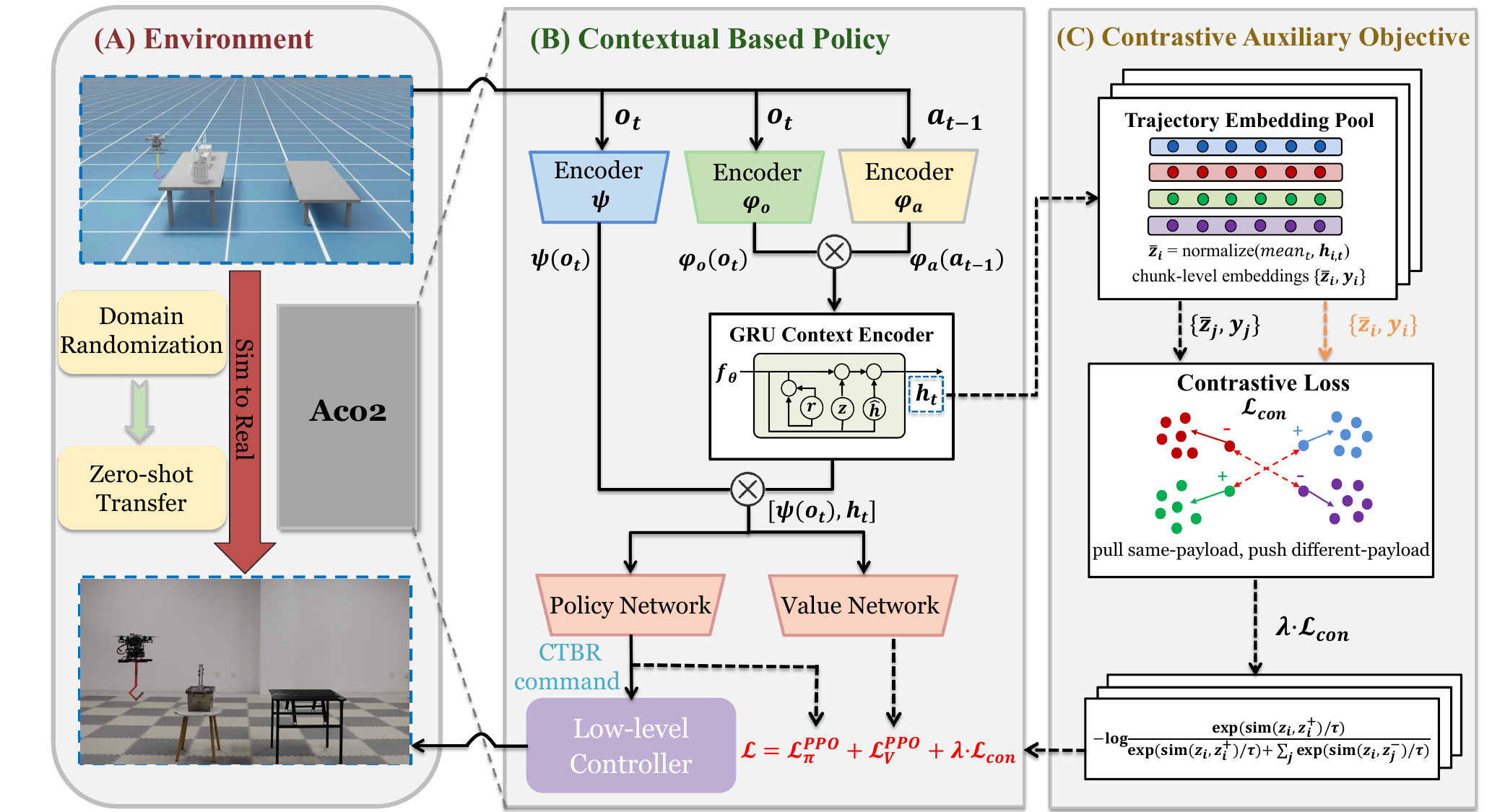} 
  \caption{The overall framework of \textit{Aco2}. }
  \label{fig:model-architecture}
\end{figure}

This section presents the methodology of the proposed \textit{Aco2} framework, with the overall architecture shown in Fig.~\ref{fig:model-architecture}. \textit{Aco2} builds on a context-based policy that infers a compact latent embedding from interaction history for online dynamics adaptation, as described in Sec.~\ref{subsec:context-policy}. Sec.~\ref{subsec:Contrastive-Context-Encoder} introduces a contrastive auxiliary objective that further structures this embedding by separating representations of different payloads, mitigating representation collapse. Finally, since learning the complete hooking, transport, and detachment sequence end-to-end is difficult, Sec.~\ref{subsec:Training-Procedure} describes the training procedure used to make optimization tractable.
\subsection{Context-Based Policy}
\label{subsec:context-policy}
\paragraph{Overall architecture}
The policy and critic architectures are shown in Fig.~\ref{fig:model-architecture}(B).  At each timestep, the current observation $o_t$ and previous action $a_{t-1}$ are first encoded by separate MLP encoders and then concatenated as the input to a GRU context encoder, which produces a history-dependent context feature. In parallel, $o_t$ is encoded by another MLP observation encoder with the same architecture as the observation encoder, producing an instantaneous state feature. The context feature and instantaneous state feature are concatenated and passed to the policy and value head, where the policy outputs parameters of a Beta distribution over bounded actions. The critic follows the same overall architecture but uses independent encoders, allowing the actor and critic to learn different temporal representations for control and value estimation. Detailed network specifications and key implementation details are provided in Appx.~\ref{subsec:training_details}.

\paragraph{Observations, actions and rewards}
At timestep $t$, the state of the UAV is given by
\(
\bm{x}_{u,t}
=
[
\bm{p}_{u,t},
\bm{q}_{u,t},
\bm{v}_{u,t},
\bm{\omega}_{u,t}
],
\)
where the subscript $t$ denotes the timestep, $\bm{p}_{u,t} \in \mathbb{R}^3$ denotes the UAV position,
$\bm{q}_{u,t} \in \mathbb{R}^4$ denotes its orientation quaternion, and
$\bm{v}_{u,t} \in \mathbb{R}^3$, $\bm{\omega}_{u,t} \in \mathbb{R}^3$ denote its linear and angular velocities, respectively.
The state of the bar attached to the hook is represented as
\(
\bm{x}_{b,t}
=
[
\bm{p}_{b,t},
\bm{q}_{b,t},
\bm{v}_{b,t},
\bm{\omega}_{b,t}
]
\)
. We also include the previous action
$\bm{a}_{t-1}$ in the observation to improve action smoothness.
To make the policy directly aware of the manipulation objective, we construct task-relative geometric observations. Let $\bm{p}_{h,t}$, $\bm{p}_{l,t}$, $\bm{p}_{o,t}$, $\bm{p}_{g,t}$, and $\bm{p}_{e,t}$ denote the positions of the hook tip, selected handle, selected object, target goal, and exit point, respectively.
The task observation is defined as
\(
\bm{x}_{r,t}
=
[
\bm{d}_{a,t},
\bm{p}_{l,t},
\bm{p}_{h,t},
\bm{p}_{g,t},
\bm{p}_{e,t},
\bm{p}_{o,t},
\bm{d}_{hl,t},
\bm{d}_{og,t},
\bm{d}_{he,t}
],
\)
where $\bm{d}_{a,t}$ contains the relative vectors from the hook tip to all candidate handles, and
\(
\bm{d}_{hl,t}
=
\bm{p}_{l,t}
-
\bm{p}_{h,t},
\quad
\bm{d}_{og,t}
=
\bm{p}_{g,t}
-
\bm{p}_{o,t},
\quad
\bm{d}_{he,t}
=
\bm{p}_{e,t}
-
\bm{p}_{h,t}.
\)
The full policy observation is then
\[
\bm{o}_t
=
[
\bm{x}_{u,t},
\bm{a}_{t-1},
\bm{x}_{b,t},
\bm{x}_{r,t}
].
\]

We use collective thrust and body-rate control (CTBR) as the action interface, where the policy specifies collective thrust and desired body rates. The policy outputs a four-dimensional action:
\[
\bm{a}_t =
\left[
c_t,\ \omega^x_t,\ \omega^y_t,\ \omega^z_t
\right],
\]
where $c_t$ denotes the normalized collective thrust command, and $\omega^x_t$, $\omega^y_t$, and $\omega^z_t$ denote the desired body rates around the roll, pitch, and yaw axes.

The reward at time $t$, denoted as $r_t$, is defined as
\[
r_t =
r^{\mathrm{hook,fine}}_t
+
r^{\mathrm{hook,coarse}}_t
+
r^{\mathrm{obj}}_t
+
r^{\mathrm{exit}}_t
-
c^{\mathrm{act}}_t
-
c^{\mathrm{vel}}_t
-
c^{\mathrm{tilt}}_t .
\]
Here $r^{\mathrm{hook,fine}}_t$ and $r^{\mathrm{hook,coarse}}_t$ encourage the hook tip to approach the selected handle. The fine-grained term provides high-precision guidance near the handle, which is necessary for successful hooking, while the coarse-grained term provides a smoother long-range signal when the hook is still far from the handle. The term $r^{\mathrm{obj}}_t$ encourages the selected object to move toward the target position, and $r^{\mathrm{exit}}_t$ guides the hook toward a predefined exit point after the object reaches the target region.

The remaining terms are regularization costs: $c^{\mathrm{act}}_t$ penalizes abrupt action changes, $c^{\mathrm{vel}}_t$ penalizes excessive velocity, and $c^{\mathrm{tilt}}_t$ penalizes unsafe tilt angles. Instead of continuously encouraging smaller action changes or lower velocities, these costs are activated only when the corresponding quantity exceeds a predefined threshold. This threshold-based design avoids conflicting with task completion while still discouraging unstable behaviors. Detailed reward formulations are provided in Appx.~\ref{subsubsec:reward_function_details}.

  \subsection{Contrastive Context Encoder}
  \label{subsec:Contrastive-Context-Encoder}
  The GRU context encoder described above produces a latent embedding at each
  timestep, but without explicit supervision there is no guarantee that these
  embeddings capture task-relevant distinctions across episodes with different
  payload contexts.
  Contrastive representation learning addresses this by shaping an embedding
  space in which trajectory segments collected under the same latent context are
   pulled together, while segments from different contexts are pushed apart.
  Given an encoder $g_\psi(\cdot)$ that maps an input $x_i$ to an embedding
  $z_i=g_\psi(x_i)$, contrastive learning constructs a positive sample $x_i^+$
  drawn from the same latent context as $x_i$, and a set of negative samples
  $\{x_j^-\}$ drawn from episodes with different contexts. A widely used
  objective is the InfoNCE loss:
  \[
  \mathcal{L}_{\mathrm{NCE}}(i)
  =
  -\log
  \frac{
  \exp(\mathrm{sim}(z_i,z_i^+)/\tau)
  }{
  \exp(\mathrm{sim}(z_i,z_i^+)/\tau)
  +
  \sum_{j}
  \exp(\mathrm{sim}(z_i,z_j^-)/\tau)
  },
  \]
  where $\mathrm{sim}(\cdot,\cdot)$ is typically cosine similarity and $\tau>0$
  is a temperature parameter. Minimizing $\mathcal{L}_{\mathrm{NCE}}$ aligns the
   embeddings of trajectory segments that share an underlying context and
  separates those that do not, so that the recurrent state produced by $g_\psi$
  becomes a discriminative summary of the latent payload context, which the
  downstream actor and critic can condition on.

In our implementation, as illustrated in Fig.~\ref{fig:model-architecture}(C), we adopt the supervised contrastive loss~\citep{khosla2020supervised} to encourage the recurrent encoder to distinguish different payload contexts. Concretely,
during PPO training, we sample a mini-batch of trajectory chunks from the replay buffer. Each chunk is
associated with a task label $y_i$ corresponding to the latent context of the episode. For the $i$-th
chunk, the valid recurrent features are averaged to obtain a context embedding:
\[
\bm{z}_i
=
\frac{1}{|\mathcal{T}_i|}
\sum_{t\in\mathcal{T}_i}
\bm{h}_{i,t},
\qquad
\bar{\bm{z}}_i
=
\frac{\bm{z}_i}{\|\bm{z}_i\|_2},
\]
where $\mathcal{T}_i$ denotes the valid timesteps before an episode boundary.

For each anchor $i$, the positive set is
\(
P(i)=\{p\mid y_p=y_i,\ p\neq i\}.
\)
Anchors without positive samples are ignored. The contrastive loss is defined as
\[
\mathcal{L}_{\mathrm{con}}
=
-\frac{1}{|\mathcal{I}|}
\sum_{i\in\mathcal{I}}
\frac{1}{|P(i)|}
\sum_{p\in P(i)}
\log
\frac{
\exp\left(\bar{\bm{z}}_i^\top \bar{\bm{z}}_p/\tau\right)
}{
\sum_{a\neq i}
\exp\left(\bar{\bm{z}}_i^\top \bar{\bm{z}}_a/\tau\right)
},
\]
where $\mathcal{I}=\{i\mid |P(i)|>0\}$ and $\tau$ is the temperature parameter.

\subsection{Training Procedure }
\label{subsec:Training-Procedure}

  \textit{Aco2} is trained with a curriculum that progressively decomposes
  the manipulation task, realized entirely by gating individual terms of
  the per-timestep reward in Sec.~\ref{subsec:context-policy} with
  indicator functions of the training step and task progress.
  Let $n$ denote the current training step, $N$ a predefined switching
  step, and $l_{o,t}$ the distance between the manipulated object and
  the target region at timestep $t$, and $\delta$ the distance threshold for considering the object delivered. The effective training reward is
\[
\tilde r_t =
r^{\mathrm{hook,fine}}_t
+ r^{\mathrm{hook,coarse}}_t
+ r^{\mathrm{obj}}_t
+ \mathbb{I}(n>N)\,\mathbb{I}(l_{o,t}<\delta)\, r^{\mathrm{exit}}_t
- \mathbb{I}(n>N)\bigl(c^{\mathrm{act}}_t + c^{\mathrm{vel}}_t + c^{\mathrm{tilt}}_t\bigr).
\]
  In the early stage ($n\!\le\!N$), both the exit reward and the
  regularization costs are switched off, letting the policy focus on
  engaging the target handle and transporting the object with
  sufficiently large control inputs. Once $n\!>\!N$, the regularization
  costs are activated to enforce smoother actions and limit excessive linear velocity and tilt angle, while the exit reward is further gated by
  $\mathbb{I}(l_{o,t}<\delta)$ and only kicks in after stable delivery,
  guiding the hook toward the predefined exit point.

  The overall training procedure follows an on-policy recurrent PPO framework. In each iteration, Aco2 collects parallel rollouts under randomized task and physical conditions, with domain randomization such as observation noise, command latency, and external torque disturbances applied to bridge the sim-to-real gap.
 The collected transitions are stored together with recurrent states and task labels. Training batches are constructed using a sliding-window strategy along the temporal dimension to improve sample efficiency. The complete training pipeline is provided in Appx.~\ref{subsec:pseudo_code_for_aco2}

\section{Experiments and Results}
\label{sec:result}
\subsection{Experimental Setup}
\label{subsec:Experimental-Setup}
We train Aco2 entirely in NVIDIA Isaac Sim~\citep{NVIDIA_Isaac_Sim} and directly deploy the learned policy to the physical platform without real-world fine-tuning. The simulation environment includes a hook-equipped quadrotor and four payload categories: medical kit, handbag, weight, and bucket. These objects differ in geometry, mass, and center of mass, inducing distinct flight dynamics after attachment and providing diverse task contexts for policy training.

We then evaluate zero-shot transfer in the real world using three container types with different geometries and randomized internal loads. Pickup and target locations vary across trials to test generalization beyond the training payloads and spatial configurations. The total payload mass ranges from 0.46 kg to 0.90 kg, covering a wide range of real-world payload conditions and approaching the practical carrying limit of the platform. Details of the experimental setup are provided in Appx.~\ref{subsec:experimental_setup}.

 \begin{table}[t]
    \centering
    \caption{Task success rates (\%) under in-distribution and out-of-distribution (OOD) settings. Results are reported as mean $\pm$ standard error across five seeds. $\Delta$ denotes the absolute improvement of w/ contrastive loss (CL) over w/o CL, measured in percentage points.}
    \label{tab:task_success}
    
    \vspace{0.25em}
    \small
    \setlength{\tabcolsep}{4.5pt}
    \renewcommand{\arraystretch}{1.0}

    \begin{tabular}{@{}llccccc@{}}
        \toprule
        Setting & Method & Med. Kit & Handbag & Weight & Bucket & Avg. \\
        \midrule

        \multirow{3}{*}{\shortstack[l]{In distribution}}
            & w/o CL   & 59.1 $\pm$ 24.1 & 29.1 $\pm$ 17.8 & 57.5 $\pm$ 23.6 & 54.7 $\pm$ 22.4 & 50.1 $\pm$ 20.6 \\
            & w/ CL    & \best{77.8 $\pm$ 19.5} & \best{65.3 $\pm$ 17.0} & \best{74.1 $\pm$ 18.6} & \best{72.2 $\pm$ 19.6} & \best{72.3 $\pm$ 18.3} \\
            & $\Delta$ & \gain{18.8} & \gain{36.2} & \gain{16.6} & \gain{17.5} & \gain{22.3} \\
        
        \cmidrule(lr){1-7}
        
        \multirow{3}{*}{\shortstack[l]{Mass OOD}}
            & w/o CL   & 45.9 $\pm$ 18.9 & 18.1 $\pm$ 9.9 & 37.5 $\pm$ 16.7 & 37.8 $\pm$ 15.6 & 34.8 $\pm$ 14.4 \\
            & w/ CL    & \best{59.4 $\pm$ 15.0} & \best{34.1 $\pm$ 9.0} & \best{46.9 $\pm$ 12.6} & \best{57.2 $\pm$ 14.4} & \best{49.4 $\pm$ 12.4} \\
            & $\Delta$ & \gain{13.4} & \gain{15.9} & \gain{9.4} & \gain{19.4} & \gain{14.5} \\
        
        \cmidrule(lr){1-7}
        
        \multirow{3}{*}{\shortstack[l]{Distance OOD}}
            & w/o CL   & 35.0 $\pm$ 17.8 & 0.9 $\pm$ 0.6 & 25.0 $\pm$ 13.0 & 13.8 $\pm$ 7.0 & 18.7 $\pm$ 7.7 \\
            & w/ CL    & \best{42.5 $\pm$ 12.2} & \best{19.1 $\pm$ 6.2} & \best{46.6 $\pm$ 13.7} & \best{60.0 $\pm$ 18.0} & \best{42.0 $\pm$ 10.6} \\
            & $\Delta$ & \gain{7.5} & \gain{18.1} & \gain{21.6} & \gain{46.2} & \gain{23.4} \\

        \bottomrule
    \end{tabular}
\end{table}
\begin{figure}[htbp]
    \centering
    \includegraphics[width=0.84\textwidth]{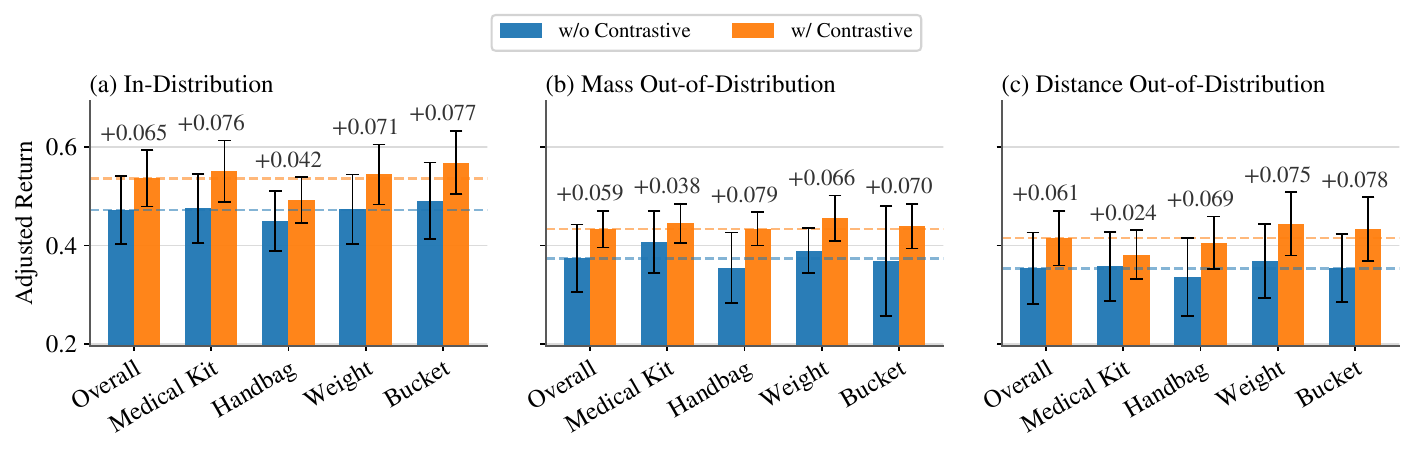}
    \vspace{-0.6em}
    \caption{Simulation return under in-distribution and two out-of-distribution settings. The return is adjusted by subtracting the mean reward at the initial environment state. Results are reported as mean $\pm$ standard error across five seeds. }
    \label{fig:bar_reward}
\end{figure}
\subsection{Simulation Evaluation}
\label{subsec:Simulation-Evaluation}
    

We evaluate task success rates under in-distribution and two out-of-distribution (OOD) settings in Table~\ref{tab:task_success}. The first OOD setting randomizes object masses outside the training range, while the second shifts the target placement location. Across all three settings, the contrastive variant consistently outperforms the baseline, confirming that the contrastive objective produces more effective context representations for adaptive control. Without the contrastive loss, performance degrades sharply under OOD conditions, especially in the distance shift setting. The contrastive variant degrades more gracefully, demonstrating stronger robustness to conditions unseen during training.

Fig.~\ref{fig:bar_reward} presents the return across all three settings. The contrastive variant achieves uniformly higher return on every payload, and the improvement remains consistent from in-distribution to both OOD conditions. Notably, the variant without contrastive loss shows growing standard error as the setting becomes more challenging, with individual payloads such as bucket exhibiting particularly large variance in the mass out-of-distribution setting, suggesting that different seeds overfit to different payload subsets. The contrastive variant, by contrast, maintains both higher mean return and tighter standard error across most payloads, indicating that the contrastive objective encourages a balanced, payload-agnostic representation rather than one that favors specific object types.

\begin{wrapfigure}[10]{r}{0.38\linewidth}
	\centering
	\vspace{-7mm}
\includegraphics[width=1\linewidth]{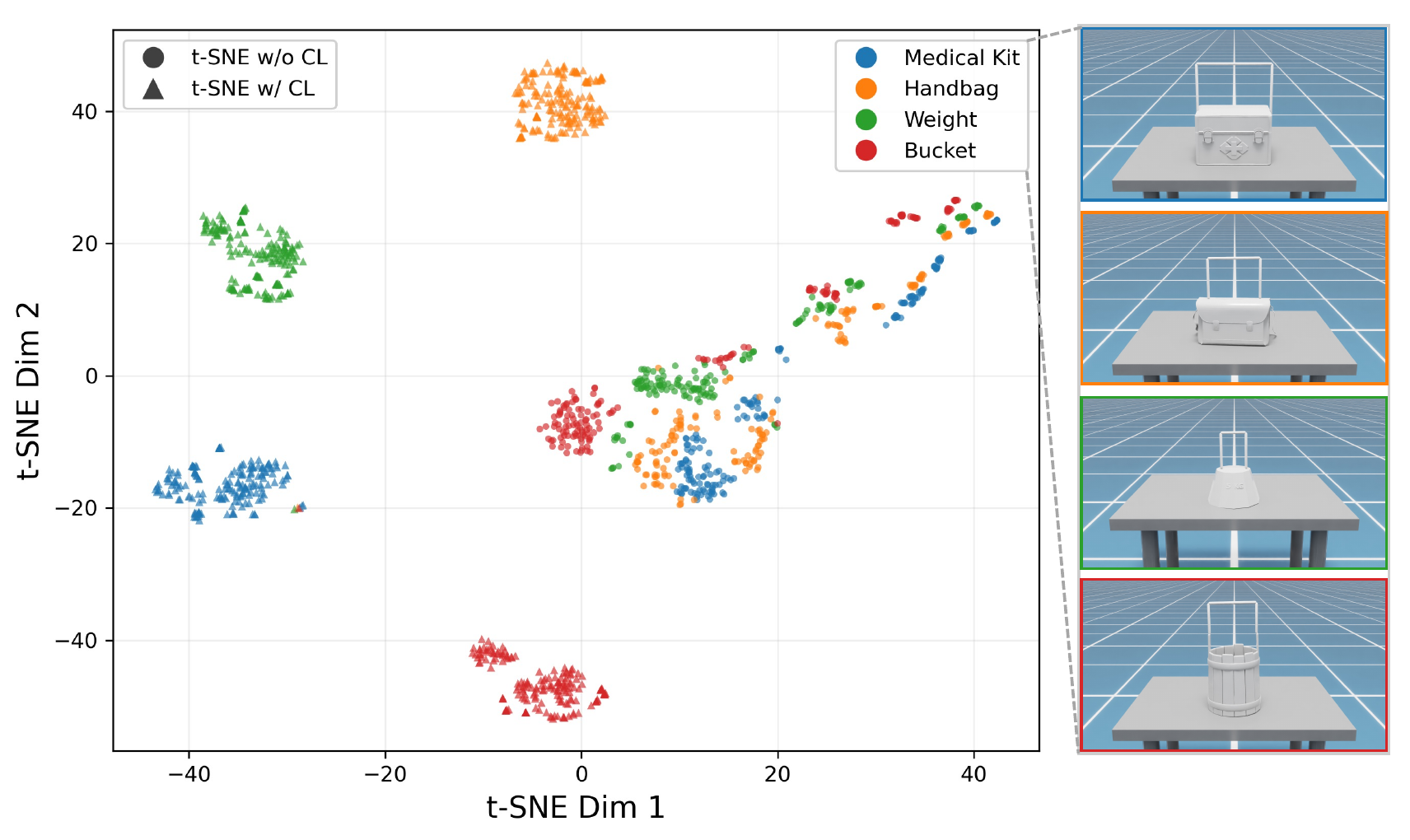}
	\vspace{-6.5mm}
	\caption{t-SNE visualization with and without contrastive loss (CL).} 
	\label{fig:tsne}
\end{wrapfigure}
We dive into the learned representations by visualizing the latent embeddings via t-SNE~\citep{van2008visualizing} in Fig.~\ref{fig:tsne}. With the contrastive loss, embeddings of different payloads form clearly separated clusters, whereas without it the representations largely overlap. This suggests that the contrastive objective mitigates representation collapse and encourages discriminative context encodings, which in turn supports effective online adaptation across diverse payloads. Additional training curves and representative simulation rollouts are provided in Appendix~\ref{subsec:simulation_training_results}, further showing that the learned policy achieves more stable convergence and adapts to payloads with distinct mass, geometry, and center-of-mass properties.

\subsection{Real-World Evaluation} 
\label{subsec:Real-World-Evaluation} 
\begin{figure}[t!]
  \centering

    \includegraphics[width=0.4\textwidth]{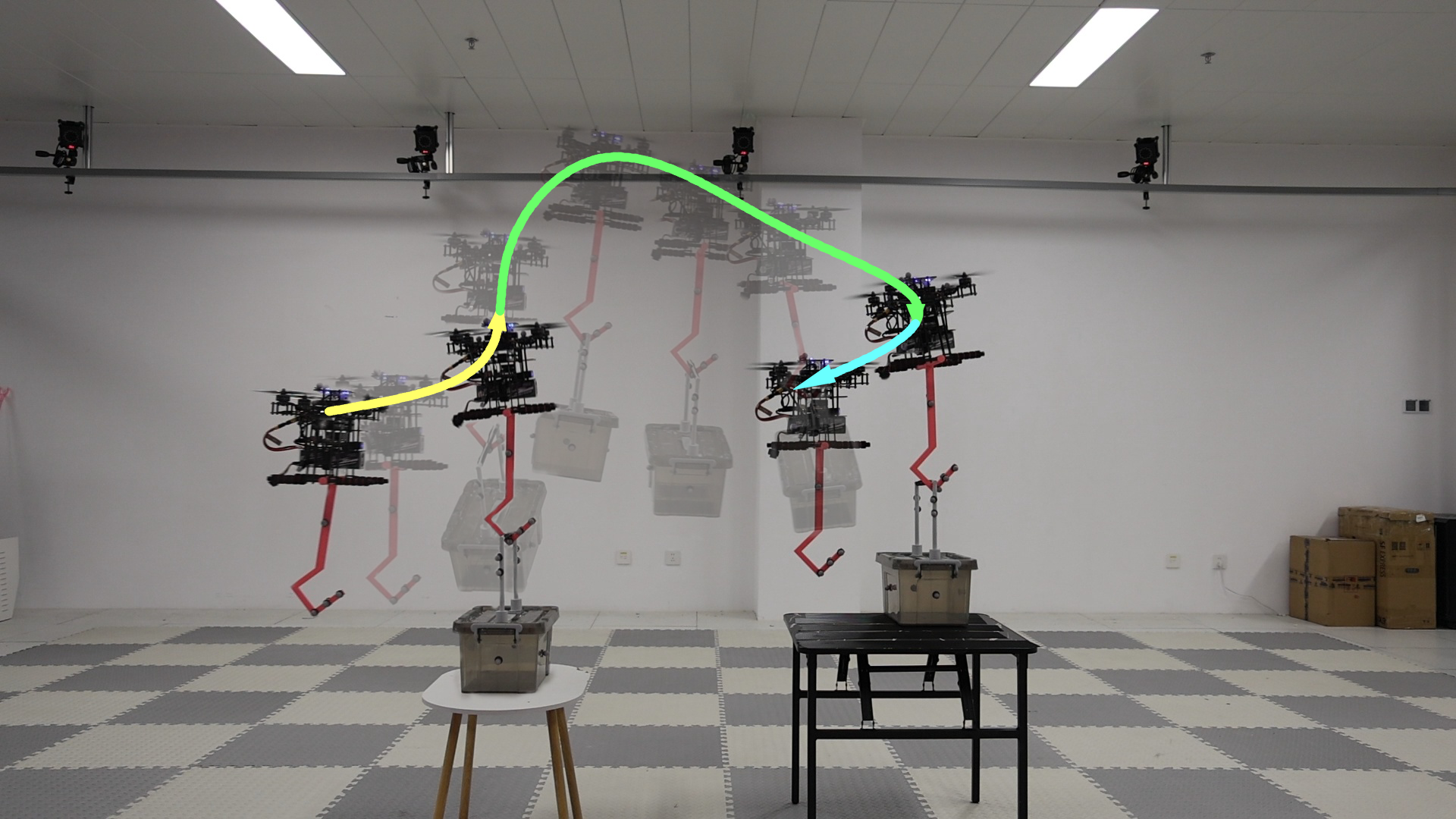}
  \hfill 
    \raisebox{-11.8pt}{\includegraphics[width=0.59\textwidth]{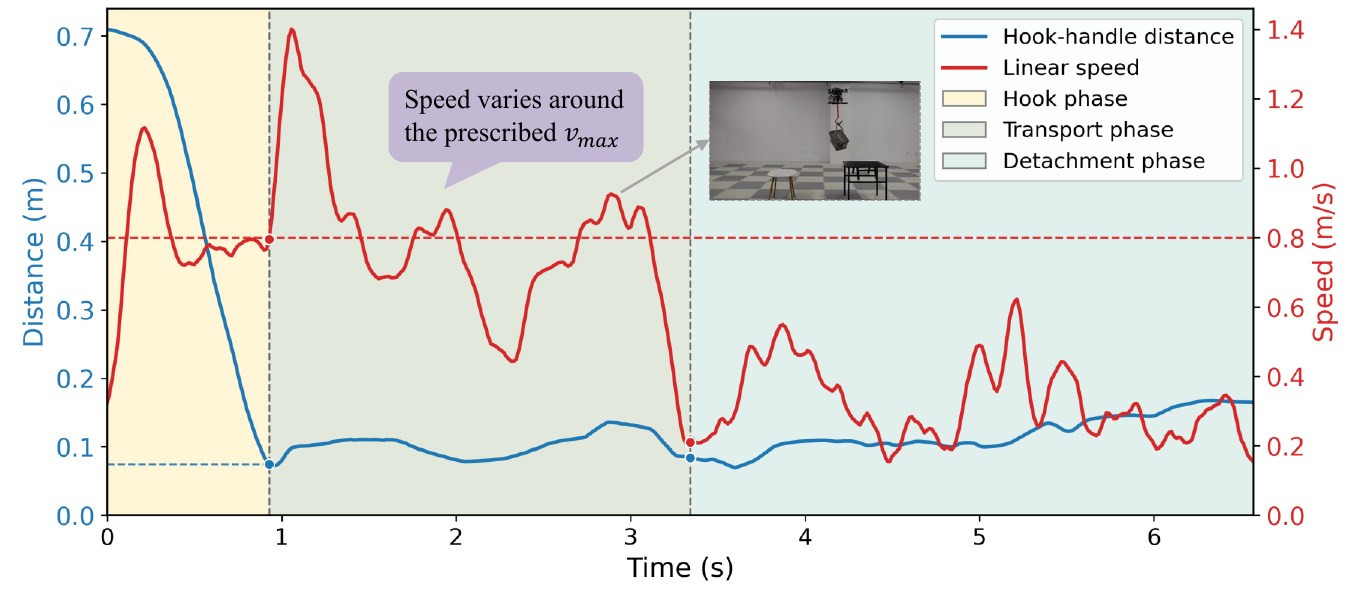}}
    \vspace{-4pt} 
    \caption{Real-world autonomous delivery trial. Left: visualization of the hook-equipped quadrotor completing an unattended delivery sequence from object acquisition to target placement. Right: analysis of the hook-handle distance and quadrotor linear speed for the trial shown on the left.}
    \label{fig:real_world_analysis}

\end{figure}

We conduct fully autonomous delivery trials by directly deploying the simulation-trained policy without any fine-tuning. Fig.~\ref{fig:real_world_analysis} shows a representative trial in which the quadrotor delivers an empty box seen during training. The quadrotor autonomously completes the full hooking, transport, and detachment sequence, with the right panel recording the hook-handle distance and linear speed throughout the three phases. During the approach for hooking and the initial transport stage, the quadrotor accelerates toward the target, while slowing down for the precise alignment required during the hooking and detachment maneuvers.

\begin{figure}[t!]
    \centering
\includegraphics[width=0.95\textwidth]{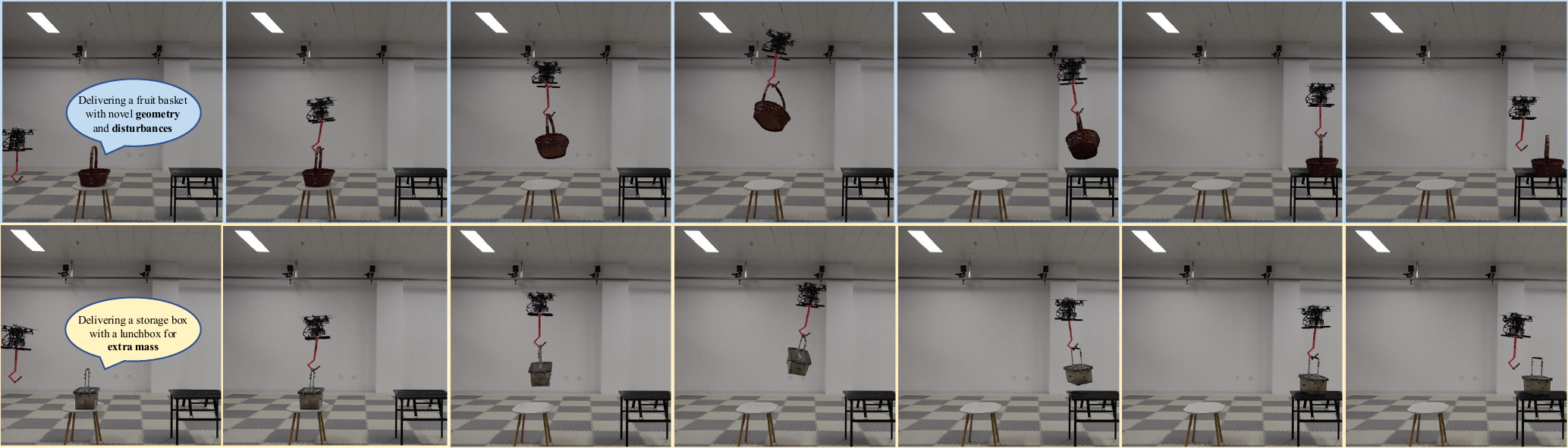}
    \caption{Generalization to unseen payloads. Top: delivery of a basket with a different geometry from the training containers. Bottom: delivery of a box loaded with an everyday meal}
    \label{fig:basket_and_empty_box}
\end{figure}

To further validate generalization, we test the policy on conditions that are difficult to capture in our rigid-body Isaac Sim environment. As shown in Fig.~\ref{fig:basket_and_empty_box}, we first replace the box with a fruit basket whose irregular geometry introduces aerodynamic disturbances absent from training. Despite this unseen payload shape, the quadrotor still completes the full delivery sequence, indicating that the learned context embedding can adapt online to altered external interactions.

We then load the box with an everyday meal, raising the total payload to 0.86 kg. Unlike the rigid objects modeled in simulation, the meal can shift inside the container during flight, continuously changing the center of mass and moment of inertia. The policy again completes the delivery without retraining or parameter adjustment, demonstrating robustness to payload dynamics that were not explicitly modeled during training.

\subsection{Ablation Study}
\label{subsec:Ablation-in-Real-World-Experiments}

To verify the necessity of each regularization term for stable physical flight, we ablate the thrust smoothness, velocity bound, angular velocity smoothness, and tilt angle penalty individually. 

\begin{wrapfigure}[10]{r}{0.54\linewidth}
	\centering
	\vspace{-6mm}
\includegraphics[width=1\linewidth]{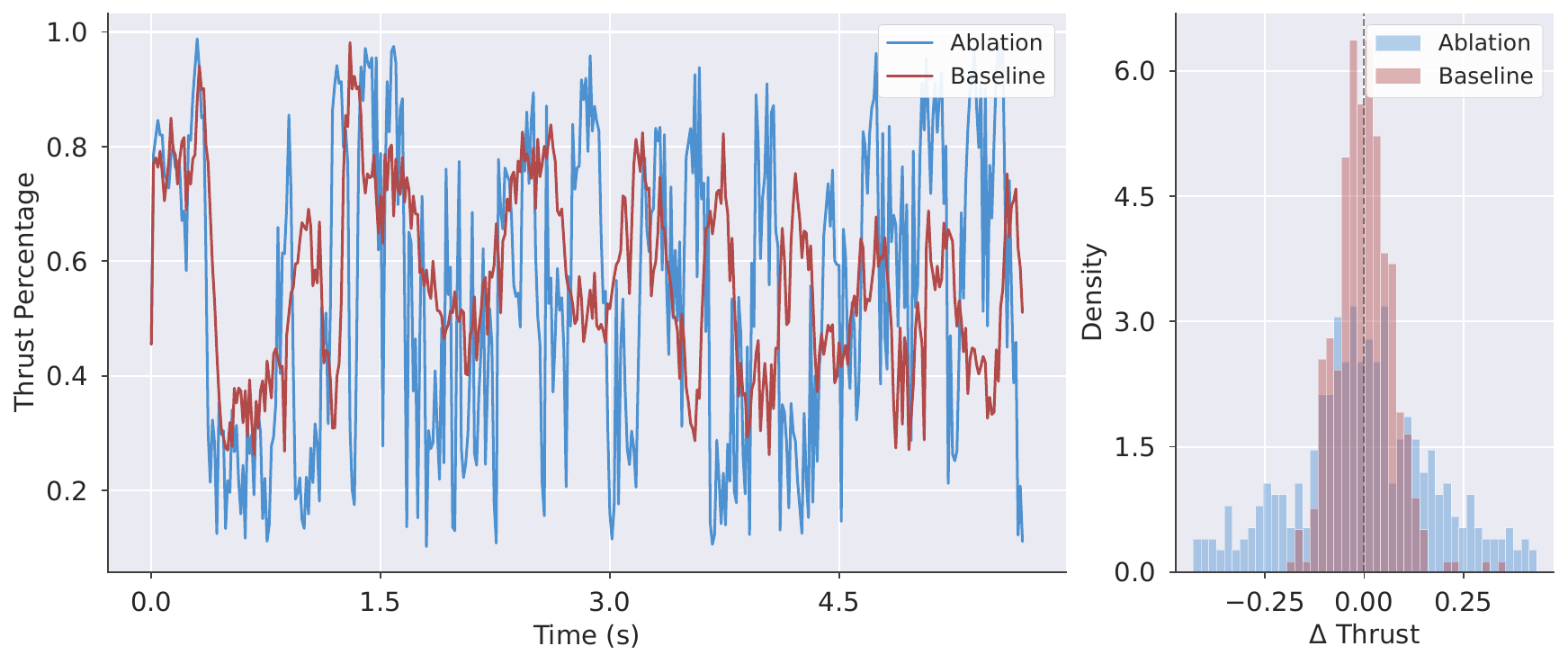}
	\vspace{-4.5mm}
	\caption{Thrust profiles w/ and w/o smoothness regularization.} 
	\label{fig:thrust_ablation_comparison}
\end{wrapfigure}
\textbf{Thrust smoothness}. Fig.~\ref{fig:thrust_ablation_comparison} compares the thrust profile and thrust change distribution with and without the smoothness regularization. With the regularization, the thrust signal varies gradually and the distribution of $\Delta$thrust is concentrated around small values. Removing it leads to noticeably more abrupt thrust fluctuations, with the distribution exhibiting a heavier tail toward large changes. While the vehicle still completes the task in this case, the aggressive thrust commands increase mechanical stress and reduce flight robustness.

In addition to the real-world ablations, we conduct sensitivity analyses on the threshold-based reward in simulation, confirming that the threshold mechanism yields more stable convergence and lower sensitivity to reward hyperparameters. Results for velocity bound, angular velocity smoothness, tilt angle regularization, and the threshold reward sensitivity analysis are all provided in Appx.~\ref{subsec:missing_ablation_study}.

  
  

\section{Conclusion}
\label{sec:conclusion}
To address the autonomous delivery problem across diverse objects, we present Autonomous Aerial Manipulation via Contextual Contrastive Meta Reinforcement Learning (\textit{Aco2}), a framework enabling a quadrotor equipped with a lightweight hook to autonomously transport diverse handle-equipped objects. At its core, a contextual encoder infers payload-induced dynamics from interaction history, while a contrastive objective sharpens this representation and strengthens generalization across payloads. We validate \textit{Aco2} in simulation and further demonstrate its effectiveness on a physical quadrotor delivering objects with substantially different shapes and masses, confirming the practical value of meta-reinforcement learning for adaptive aerial transport.
\section{Limitations}
\label{sec:limitations}
Our system relies on external motion capture for state estimation, restricting experiments to controlled lab settings. Future work could replace it with onboard visual-inertial odometry and handle detection, explore end-to-end visuomotor policies that directly map onboard camera observations to control actions for deployment in less instrumented environments. In addition, the payload capacity of a single quadrotor limits the mass of objects \textit{Aco2} can engage and
  transport, so sufficiently heavy targets remain out of reach.
  Extending \textit{Aco2} to multi-UAV cooperation would lift this
  ceiling~\citep{yuan2023survey,feng2026multi}, but requires handling load sharing, inter-robot coordination,
  communication constraints, and coupled dynamics, motivating
  decentralized or shared-context policies for larger payloads.
\clearpage
\newpage

\bibliographystyle{ACM-Reference-Format}
\bibliography{aaai24}

\newpage
\onecolumn
\appendix

\setcounter{secnumdepth}{3} 
\section{Experimental Details}
\label{sec:experimental_details}
\subsection{Experimental Setup}
\label{subsec:experimental_setup}
\begin{figure}[htbp]
    \centering
    \includegraphics[width=0.9\textwidth]{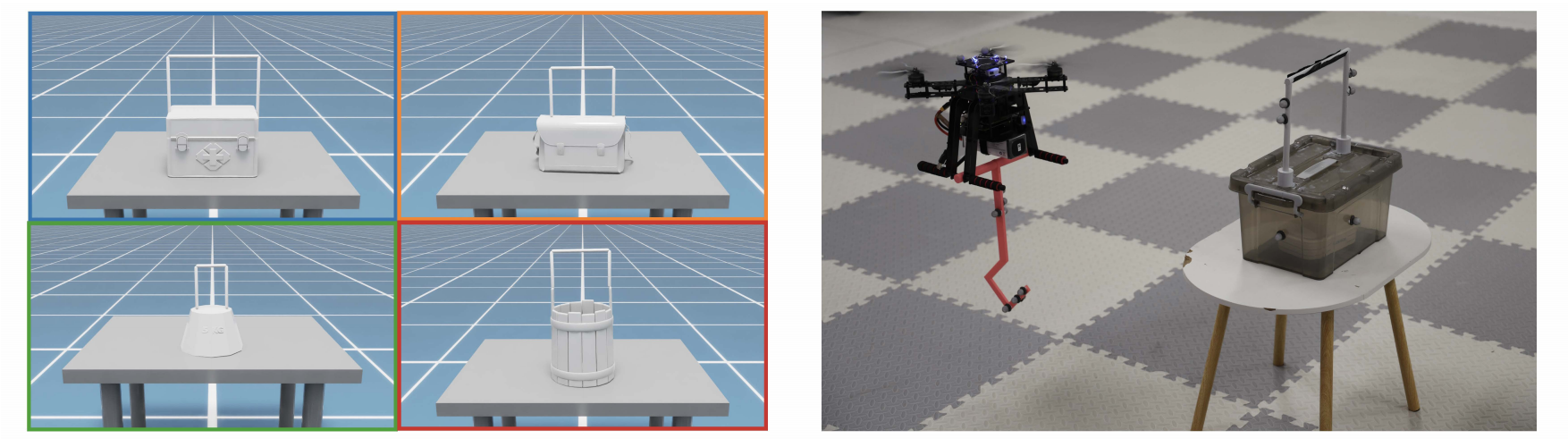}
    \caption{Simulation and real-world experimental setups}
    \label{fig:experimental_setup}
\end{figure}

This section details the simulation and real-world experimental settings. Representative payloads and the physical deployment setups are shown in Fig.~\ref{fig:experimental_setup}.
\subsubsection{Simulation Setup}
\label{subsubsec:simulation-setup}
All simulation experiments are conducted in NVIDIA Isaac Sim 5.0.0 with 2048 parallel simulation environments on a single NVIDIA RTX 5090 GPU. The simulated environment contains a hook-equipped quadrotor and four payload categories: medical kit, handbag, weight, and bucket. These objects are designed to induce distinct payload dynamics through variations in mass distribution, total mass, and geometry, thereby providing diverse task contexts for evaluating online adaptation.

The medical kit and handbag have similar overall object scales but differ in their center-of-mass configurations. Specifically, we shift their centers of mass in opposite directions in the object frame, modeling asymmetric internal loading that commonly occurs in real containers. To further approximate the rotational disturbances observed during real flight, we additionally apply external torques to the quadrotor body in the corresponding directions. This compensates for aerodynamic effects that are difficult to model accurately in simulation, such as downwash-induced payload motion and its feedback on the vehicle.
The weight is designed to be significantly heavier than the other payloads, with a mass approximately three times larger than the remaining object categories. This setting evaluates whether the policy can adapt to large changes in payload-induced inertia and required thrust. In contrast, the bucket is characterized primarily by its large volume, which introduces different geometric constraints during hooking and transport despite not being the heaviest object. Together, these four payload categories cover complementary sources of variation, including center-of-mass offset, payload mass, and object size.

\subsubsection{Real-world Setup}
\label{subsubsec:real-world-setup}
For real-world deployment, we use a quadrotor with a diagonal size of 290 mm and a total takeoff weight of 2.09 kg. The learned policy is executed onboard an NVIDIA Jetson Orin NX and outputs collective thrust and body rate (CTBR) commands to the low-level flight controller. The measured onboard inference latency is approximately 5--10 ms.

To evaluate generalization under real-world payload variation, we prepare three container types with different geometries. In each trial, a random subset of four object categories is loaded into the container, producing different internal mass distributions and total payload masses. Pickup and target locations are also randomized across trials. The total payload mass ranges from 0.46 kg to 0.90 kg, approaching the practical carrying limit of the platform.

During deployment, the policy observation is constructed from a NOKOV Motion Capture System and the onboard inertial measurement unit (IMU). The motion capture system provides the UAV position, attitude, and linear velocity, the payload bar pose and velocity, and the positions of task-relevant markers. The UAV body angular velocity is obtained from the onboard IMU, replacing the corresponding motion capture estimate for closer consistency with the actual flight controller state.

\begin{table}[t]
\centering
\caption{Domain randomization ranges.}
\label{tab:domain-random-params}
\setlength{\tabcolsep}{10pt}
\renewcommand{\arraystretch}{1.01}
\begin{tabular}{ll|ll}
\toprule
Symbol (unit) & Range & Symbol (unit) & Range \\
\midrule
$\Delta m_p$ (kg) & [-0.1, 0.1] & $\bm{s}_I$ & [0.8, 1.2]  \\
$T_{\max}$ (N) & [45.5, 46.5] & $\alpha_{\mathrm{up}}$ & [0.6, 0.85] \\
$\alpha_{\mathrm{down}}$ & [0.65, 0.9] & $\sigma_T$ (N) & 0.01 \\
$p_{\mathrm{delay}}$ & 0.5 & $d_h$ & [0.003, 0.004] \\
$\bm{\tau}_p$ (N$\cdot$m) & [-0.2, 0.2] & $\Delta c_{p,x}$ (m) & [-0.02, 0.02] \\
$\Delta c_{p,y}$ (m) & [-0.02, 0.02] & $\Delta c_{p,z}$ (m) & [-0.03, 0] \\
$\epsilon_p$ (m) & [-0.03, 0.03] & $\epsilon_q$ (deg) & [-1, 1] \\
$\epsilon_v$ (m/s) & [-0.05, 0.05] & $\epsilon_\omega$ (rad/s) & [-0.03, 0.03] \\
$\Delta p_{u,0}^{x}$ (m) & [-0.2, 0.2] & $\Delta p_{u,0}^{y}$ (m) & [-0.2, 0.2] \\
$\Delta p_{u,0}^{z}$ (m) & [0, 0.2] & $\bm{v}_{u,0}$ (m/s) & [-0.1, 0.1] \\
$\omega_{u,0}^{x}$ (deg/s) & [-0.3, 0.3] & $\omega_{u,0}^{y}$ (deg/s) & [-0.3, 0.3] \\
$k_d$ & [0.1, 0.3] &  &  \\
\bottomrule
\end{tabular}
\end{table}
\subsection{Domain Randomization}
\label{subsubsec:domain_randomization}
In our domain randomization setup for unmanned aerial vehicles, several parameters are adjusted to enhance the
robustness and generalization of the policy. The corresponding randomization ranges are reported in Table~\ref{tab:domain-random-params}.
\begin{itemize}
\item \texttt{payload\_mass\_range}: Perturbs the payload mass by $\Delta m_p$ around the nominal value.

    \item \texttt{uav\_inertia\_range}: Scales the quadrotor inertia tensor by $\bm{s}_I$, changing the quadrotor's angular response to control commands.

    \item \texttt{max\_thrust\_range}: Changes the maximum motor thrust $T_{\max}$ to model voltage-dependent thrust variation and actuator performance changes.

\item \texttt{motor\_delay\_coefficient\_range}: Models motor response latency by varying the current command weight $\alpha$ in
$\bm{T}_{\mathrm{applied},t}=\alpha\bm{T}_{\mathrm{cmd},t}+(1-\alpha)\bm{T}_{\mathrm{applied},t-1}$.
Separate coefficients $\alpha_{\mathrm{up}}$ and $\alpha_{\mathrm{down}}$ are sampled for increasing and decreasing thrust commands to account for asymmetric motor spin-up and spin-down responses.

\item \texttt{thrust\_noise\_range}: Adds zero-mean Gaussian thrust noise to motor thrust after motor delay computation, with standard deviation $\sigma_T$ capturing unmodeled actuator uncertainty and thrust fluctuations.

\item \texttt{hook\_joint\_damping\_range}: Varies the hook joint damping coefficient $d_h$, changing how payload swing and contact impulses are transmitted from the hook to the quadrotor.

    \item \texttt{payload\_torque\_range}: Applies external body torque $\bm{\tau}_p$ to approximate payload-induced rotational disturbances.
    
\item \texttt{payload\_com\_offset\_range}: Adds a perturbation $\Delta \bm{c}_p$ to the nominal payload center-of-mass offset, with separate ranges for the horizontal components $\Delta c_{p,x}, \Delta c_{p,y}$ and the vertical component $\Delta c_{p,z}$.

\item \texttt{observation\_noise\_range}: Adds observation noise to policy inputs, with separate noise terms for position $\epsilon_p$, orientation $\epsilon_q$, linear velocity $\epsilon_v$, and angular velocity $\epsilon_\omega$.

\item \texttt{initial\_position\_range}: Varies the initial position offset of the quadrotor, with separate ranges for the horizontal components $\Delta p_{u,0}^{x}, \Delta p_{u,0}^{x}$ and the vertical component $\Delta p_{u,0}^{z}$.

\item \texttt{uav\_initial\_velocity\_range}: Varies the quadrotor initial linear velocity $\bm{v}_{u,0}$ and roll/pitch angular velocity components $\omega_{u,0}^{x}, \omega_{u,0}^{y}$, which mainly affect attitude transients.

    \item \texttt{air\_drag\_coefficient\_range}: Varies the aerodynamic drag coefficient $k_d$.


\end{itemize}

\subsection{Reward Function Details}
\label{subsubsec:reward_function_details}
We design the reward components to support both precise task execution and control behaviors suitable for hardware deployment. The task-related terms use exponential distance rewards, providing dense gradients for fine control near the handle, target position, and exit point. The control costs are defined with thresholded linear penalties, so they are activated only when action changes, velocity, or tilt exceed predefined smoothness or safety bounds. This design discourages overly conservative behavior within the normal operating range while still penalizing unsafe or unstable motions. The reward function components are formulated as:
\begin{equation}
\begin{aligned}
r^{\mathrm{hook,fine}}_t&= \lambda_1\exp{(-\lambda_2\|\bm{p}_{l,t}-\bm{p}_{h,t}\|)},\\
r^{\mathrm{hook,coarse}}_t&=\lambda_3\exp{(-\lambda_4\|\bm{p}_{l,t}-\bm{p}_{h,t}\|)},\\
r^{\mathrm{obj}}_t&=\lambda_5\exp{(-\lambda_6\|\bm{p}_{g,t}-\bm{p}_{o,t}\|)},\\
r^{\mathrm{exit}}_t&=\lambda_7\exp{(-\lambda_8\|\bm{p}_{e,t}-\bm{p}_{h,t}\|)},\\
c^{\mathrm{act}}_t&=\lambda_9\|\text{relu}(|\bm{a}_t-\bm{a}_{t-1}|-\bm{a}_{\text{threshold}})\|,\\
c^{\mathrm{vel}}_t&=\lambda_{10}\|\text{relu}(\bm{v}_{u,t}-\bm{v}_{\text{threshold}})\|,\\
c^{\mathrm{tilt}}_t&=\lambda_{11}\|\text{relu}(\bm{\theta}_{u,t}-\bm{\theta}_{\text{threshold}})\|,
\end{aligned}
\end{equation}
where $\theta_{u,t} \in [0, \pi]$ denotes the tilt angle of the quadrotor relative to the vertical $z$-axis at time $t$.
The reward parameters are scheduled with a two-stage curriculum, as summarized in Table~\ref{tab:reward-params}. In the first stage, the task is initialized with relaxed regularization to encourage exploration and successful object acquisition. In the second stage, the task difficulty is increased and the control regularization is tightened to encourage hardware-compatible behaviors, including smoother actions, lower velocity, and safer flight attitudes.
\begin{table}[t]
\centering
\caption{Reward parameters used in the two-stage curriculum. Stage I emphasizes object acquisition and transport, while Stage II activates exit guidance and control regularization for real-world deployment.}
\label{tab:reward-params}
\begin{tabular}{lccc}
\toprule
Reward term & Parameter & Stage I & Stage II \\
\midrule
Hooking reward (fine) & $\lambda_1, \lambda_2$ & 0.3, 10.0 & 0.0, 0.0 \\
Hooking reward (coarse) & $\lambda_3, \lambda_4$ & 0.3, 0.5 & 0.0, 0.0 \\
Object-to-goal reward & $\lambda_5, \lambda_6$ & 1.6, 0.5 & 1.6, 0.5 \\
Exit reward & $\lambda_7, \lambda_8$ & 0.0, 0.0 & 2.0, 1.8 \\
Action smoothness penalty & $\lambda_9, \bm{a}_{\mathrm{threshold}}$ & 0.0, 0.0 & 0.2, 0.12 \\
Velocity penalty & $\lambda_{10}, \bm{v}_{\mathrm{threshold}}$ & 0.0, 0.0 & 0.3, 0.8 \\
Tilt penalty & $\lambda_{11}, \bm{\theta}_{\mathrm{threshold}}$ & 0.0, 0.0 & 0.2, 0.17 \\
\bottomrule
\end{tabular}
\end{table}
\subsection{Training Details}
\label{subsec:training_details}
\subsubsection{Network Architecture}
\label{subsubsec:network_architecture}

The recurrent context branch encodes the current observation $o_t$ and the previous action $a_{t-1}$ with separate one-layer MLP encoders, each producing a 128-dimensional feature. This keeps observation and action information at the same representational scale before their concatenation and input to a single-layer GRU with a 128-dimensional hidden state.

The instantaneous observation branch uses a two-layer MLP encoder that maps the current observation to a 128-dimensional feature, matching the dimension of the GRU context. The policy and value networks therefore receive the feature formed by concatenating the instantaneous observation embedding and the history-dependent context.

The actor head is a two-layer MLP with 256 hidden units per layer and Tanh activations. It outputs the $\alpha$ and $\beta$ parameters of a Beta distribution for each action dimension. The critic uses an independent encoder stack with the same dimensions as the actor, followed by a two-layer MLP value head with 128 hidden units per layer and a scalar output.
\subsubsection{PPO Hyperparameters}
\label{subsubsec:ppo_hyperparameters}
The PPO optimization hyperparameters used in our experiments are summarized in Table~\ref{tab:ppo-optim-hyperparams}.
\begin{table}[t]
\centering
\caption{PPO optimization hyperparameters.}
\label{tab:ppo-optim-hyperparams}
\small
\setlength{\tabcolsep}{5pt}
\renewcommand{\arraystretch}{1.08}
\begin{tabular}{@{}ll@{\quad}ll@{}}
\toprule
Hyperparameter & Value & Hyperparameter & Value \\
\midrule
Discount factor $\gamma$ & 0.99
& Final Critic learning rate & 1e-5 \\

GAE parameter $\lambda_{\mathrm{GAE}}$ & 0.95
& RNN learning-rate multiplier & 0.03 \\

PPO clipping coefficient $\epsilon_{\mathrm{clip}}$ & 0.2
& Optimizer & Adam \\

Entropy coefficient & 0.01
& Adam epsilon & 1e-5 \\

Maximum gradient norm & 10.0
& Number of PPO epochs per update $K$ & 4 \\

Actor learning rate & 3e-4
& Mini-batch size & 1024 \\

Critic learning rate & 5e-4
& Rollout length per environment $T$ & 128 \\

Learning rate schedule & Linear decay
& Maximum episode length & 450 \\

Final Actor learning rate & 5e-6
& Recurrent training chunk $L$ & 32 \\

Curriculum switching step $N$ & $6\times10^4$
& Advantage normalization & True \\
\bottomrule
\end{tabular}
\end{table}
\subsection{Pseudo-code For \textit{Aco2}}
\label{subsec:pseudo_code_for_aco2}
\begin{algorithm}[t]
\caption{Training procedure of \textit{Aco2}}
\label{alg:aco2_training}
\begin{algorithmic}[1]
\Require Parallel environments $\mathcal{E}$, recurrent policy $\pi_\theta$, value function $V_\phi$, rollout buffer $\mathcal{B}$, rollout length $T$, chunk length $L$, PPO epochs $K$, contrastive weight $\lambda_{\mathrm{con}}$, entropy coefficient $\beta$
\State Initialize $\theta,\phi,\mathcal{B}$ and recurrent states $\bm{h}^{\pi},\bm{h}^{V}$
\For{training iteration $=1,2,\dots$}
    \State Collect $T$-step recurrent rollouts in $\mathcal{E}$ using $\pi_\theta$ with domain randomization in Appx.~\ref{subsubsec:domain_randomization}
    \State $\mathcal{B}\leftarrow \mathcal{B}\cup \{o_t,a_t,r_t,d_t,y_t,\bm{h}^{\pi}_t,\bm{h}^{V}_t\}_{t=1}^{T}$
    \State Construct overlapping trajectory chunks of length $L$ using a sliding temporal window
    \For{epoch $=1,\dots,K$}
        \State Update the recurrent actor--critic with PPO and supervised contrastive context loss:
        \[
\mathcal{L}
=
\mathcal{L}_{\mathrm{clip}}
+
 \mathcal{L}_{\mathrm{value}}
-
\beta \mathcal{H}
+
\lambda_{\mathrm{con}}\mathcal{L}_{\mathrm{con}} .
        \]
    \EndFor
    \State Clear rollout buffer $\mathcal{B}$
\EndFor
\end{algorithmic}
\end{algorithm}
 Algorithm~\ref{alg:aco2_training} outlines the training procedure of \textit{Aco2},
  which follows the curriculum introduced in Sec.~\ref{subsec:Training-Procedure}:
  once the switching step is reached, the reward formulation transitions to its
  second stage, thereby shifting the task distribution that the policy
  effectively sees. Throughout rollout collection, every parallel environment
  is reset upon hitting the maximum episode length or a termination condition,
  and the task configuration together with the domain randomized physical
  parameters is resampled at that moment. As a result, the recurrent policy is
  continually confronted with varying payloads, spatial layouts, actuation
  delays, and other sources of system variation.
  
\section{Additional Results}
\label{sec:additional_results}
\subsection{Simulation Training Results}
\label{subsec:simulation_training_results}
\begin{wrapfigure}[15]{r}{0.4\linewidth}
	\centering
	\vspace{-6mm}
\includegraphics[width=1\linewidth]{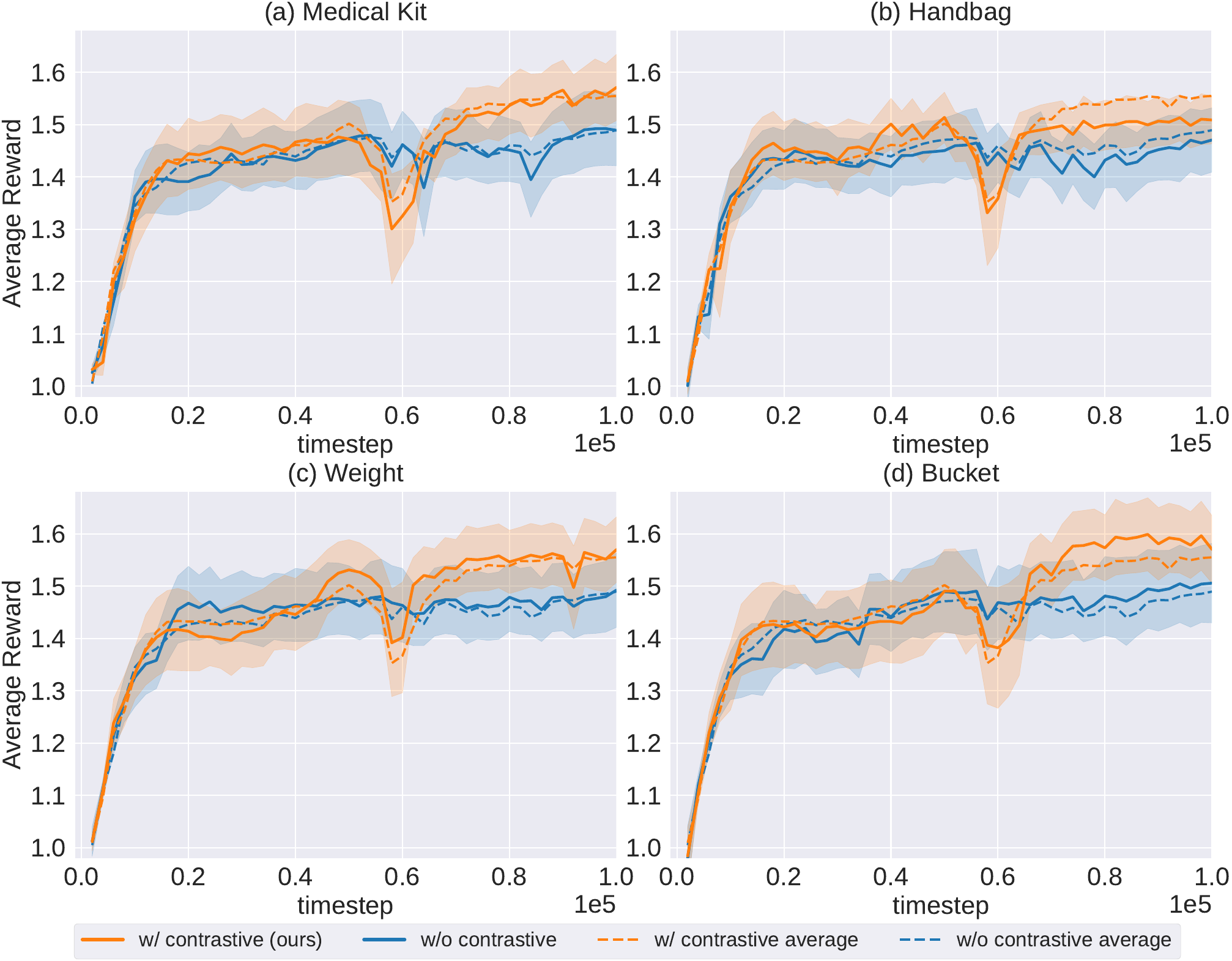}
	\vspace{-4.5mm}
	\caption{Training curves w/ and w/o contrastive loss.} 
	\label{fig:main_reward}
\end{wrapfigure}
Fig.~\ref{fig:main_reward} compares the training curves of \textit{Aco2} with and without the contrastive auxiliary loss. The evaluation reward covers only the hooking and transport phases, isolating the core adaptive control challenge from the detachment procedure. Solid lines denote the per-task reward for each payload, while dashed lines indicate the average reward across all four tasks, serving as a shared reference to reveal how each individual payload's learning progress deviates from the overall trend. The contrastive variant consistently achieves higher final reward on all payloads and exhibits more stable convergence in the later stages of training.

\begin{figure}[t!]
    \centering
    \includegraphics[width=0.9\textwidth]{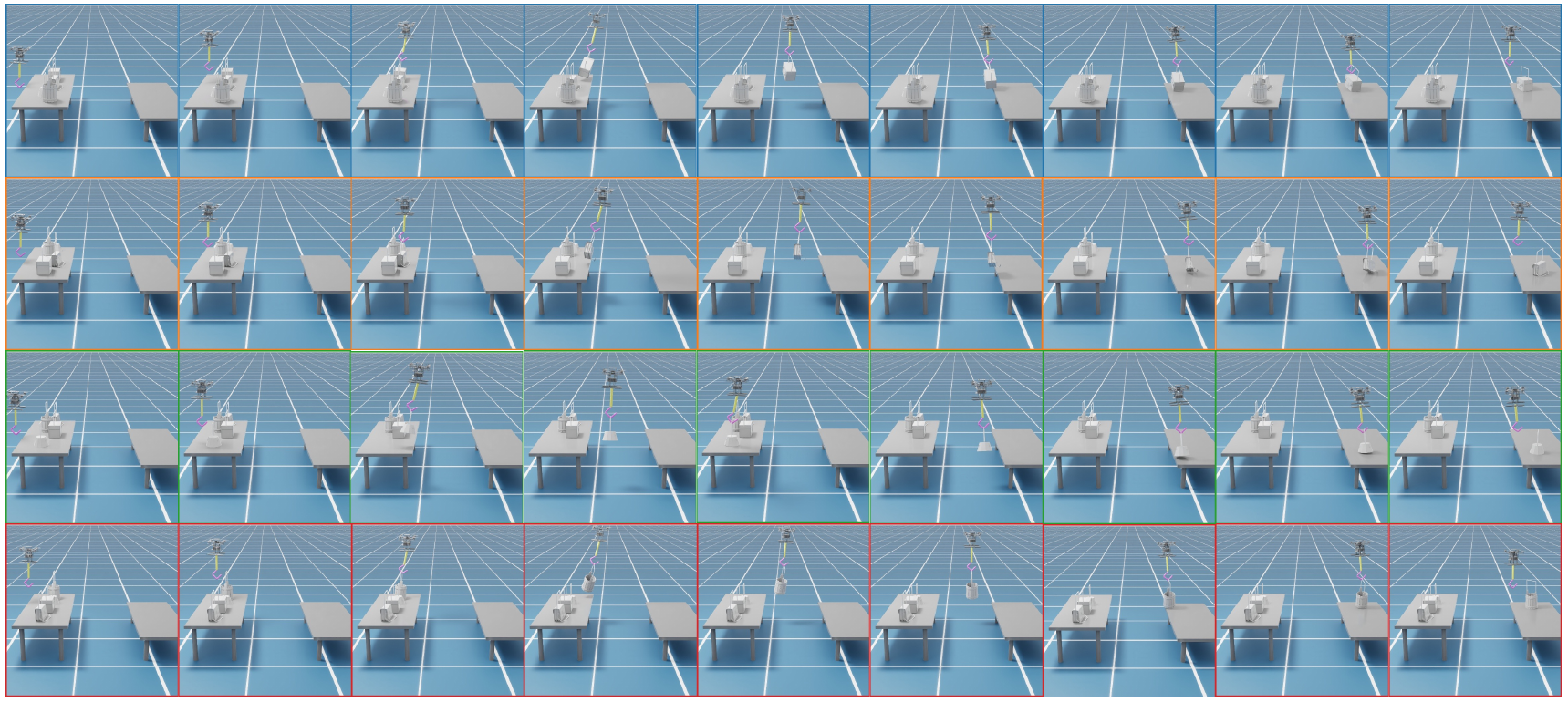}
\caption{Representative simulation rollouts across four payload categories. Rows correspond to the medical kit, handbag, weight, and bucket, respectively. The sequences illustrate successful delivery under distinct payload behaviors, including lateral sliding caused by shifted centers of mass, large mass variation, and geometric variation.}
    \label{fig:simulation_visualization}
\end{figure}

Beyond the reward curves, Fig.~\ref{fig:simulation_visualization} visualizes representative simulation rollouts for the four payload categories. For the medical kit and handbag, the effect of the shifted center of mass is directly visible during hooking and early transport: although the hook engages near the handle center, the payloads slide and tilt toward opposite sides instead of remaining centered under the hook. This produces different lateral load directions and rotational disturbances that the policy must compensate for online. For the weight and bucket, the visual difference is less about lateral sliding and more about transport response: the weight imposes a much larger load, while the bucket presents a larger geometry. The learned policy completes delivery in all cases, showing that it adapts to the distinct behaviors induced by each payload rather than relying on a single nominal transport pattern.

\subsection{Additional Ablation Studies}
\label{subsec:missing_ablation_study}
This section provides additional ablation studies on the reward design. In the main text, we report the representative real-world ablation on action smoothness regularization. Here, we provide the remaining results for the hardware-oriented regularization terms, including velocity bound, angular velocity smoothness, and tilt angle regularization. We then analyze the threshold-based reward formulation in simulation, showing how activating penalties only beyond predefined bounds affects training stability and sensitivity to reward hyperparameters.
\subsubsection{Real-World Regularization Ablations}
\label{subsubsec:real_world_regularization_ablations}
\begin{figure}[htbp]
    \centering
    \includegraphics[width=0.98\textwidth]{ 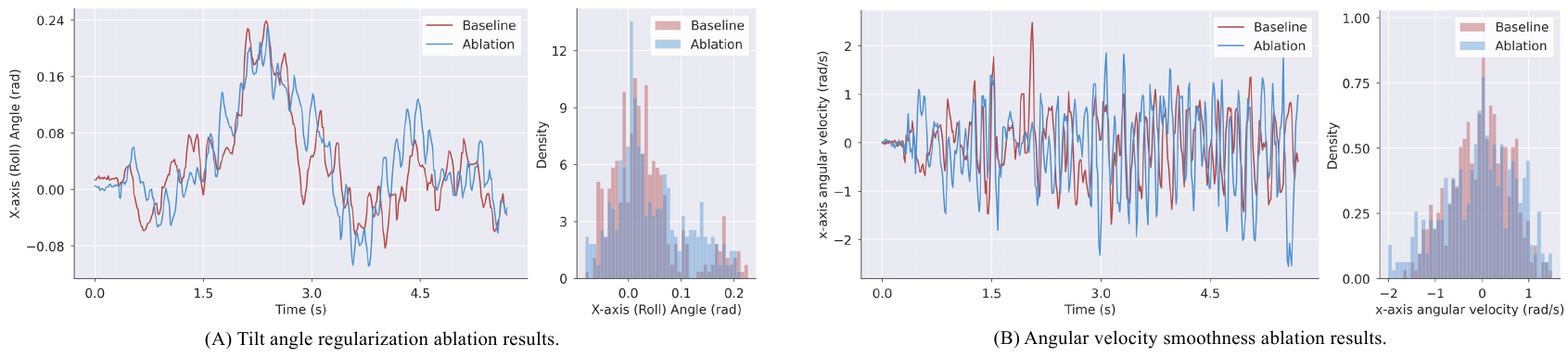}
\caption{Additional real-world ablations on attitude regularization. (A) Tilt angle regularization, showing the roll angle trajectory and distribution. (B) Angular velocity smoothness regularization, showing the $x$-axis angular velocity trajectory and distribution.}
    \label{fig:tilt_and_angular_ablation}
\end{figure}

\paragraph{Tilt angle regularization.}
We further evaluate the effect of the tilt-angle regularization in real-world deployment. Since the quadrotor primarily moves along the $x$ direction during transport, the roll angle directly reflects lateral attitude variation and flight stability. As shown in Fig.~\ref{fig:tilt_and_angular_ablation}A, the policy with tilt angle regularization keeps the roll angle concentrated within a smaller range for most of the trajectory. This indicates a more stable flight attitude, reducing the risk of side tilting or rollover when carrying a suspended payload. By maintaining a larger attitude stability margin, the regularized policy is also expected to be more robust to stronger external disturbances and payload-induced oscillations.

\begin{wrapfigure}[19]{r}{0.37\linewidth}
	\centering
	\vspace{-3mm}
\includegraphics[width=1\linewidth]{ 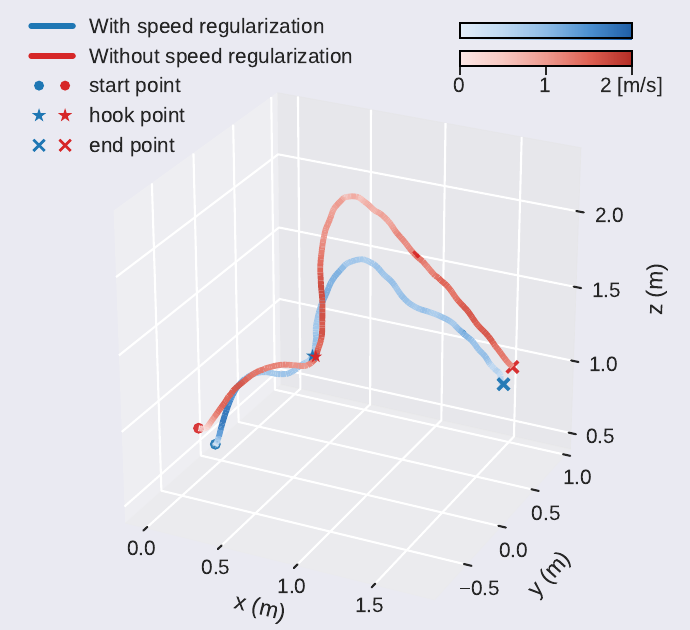}
	\vspace{-2.5mm}
\caption{Additional real-world ablation on velocity regularization, showing the 3D transport trajectories with and without the velocity bound. Trajectory color indicates the quadrotor linear speed.}
	\label{fig:speed_reg_3d_trajectory_comparison}
\end{wrapfigure}
\textbf{Velocity bound}. Fig.~\ref{fig:speed_reg_3d_trajectory_comparison} compares flight trajectories with and without velocity regularization. With the regularization, the quadrotor maintains an average speed of 0.803 m/s, close to the prescribed $v_{\max}=0.8$ m/s, whereas removing it increases the average speed to 1.097 m/s. The unregularized policy also spends 74.78\% of the trajectory above 0.9 m/s, compared with 30.37\% when the velocity bound is used, indicating that high-speed motion becomes persistent rather than occasional. This is also reflected in the spatial trajectory: without velocity regularization, the quadrotor reaches a higher transport arc with a maximum height of 2.30 m, while the regularized policy remains lower at 1.88 m. These results show that the velocity bound keeps the policy closer to the intended operating envelope instead of merely reducing the average speed. As speed increases beyond the designed range, controller vibration becomes more pronounced and the sim-to-real dynamics gap widens, since our quasi-static Isaac Sim simulation does not capture aerodynamic effects that become dominant at higher speeds.

\paragraph{Angular velocity smoothness.}
We also evaluate the regularization on angular-velocity smoothness using the UAV's $x$-axis angular velocity. As shown in Fig.~\ref{fig:tilt_and_angular_ablation}B, removing this regularization leads to large oscillations around zero, indicating frequent alternating roll rate commands. In real flight, this behavior appears as visible left-right jittering of the quadrotor and results in unstable transport. With angular-velocity smoothness regularization, the angular velocity remains more concentrated around zero, leading to smoother attitude motion and more stable payload transport.
\subsubsection{Threshold-Based Reward Sensitivity}
\label{subsubsec:threshold_reward_sensitivity}
\begin{figure}[htbp]
    \centering
    \includegraphics[width=\textwidth]{ 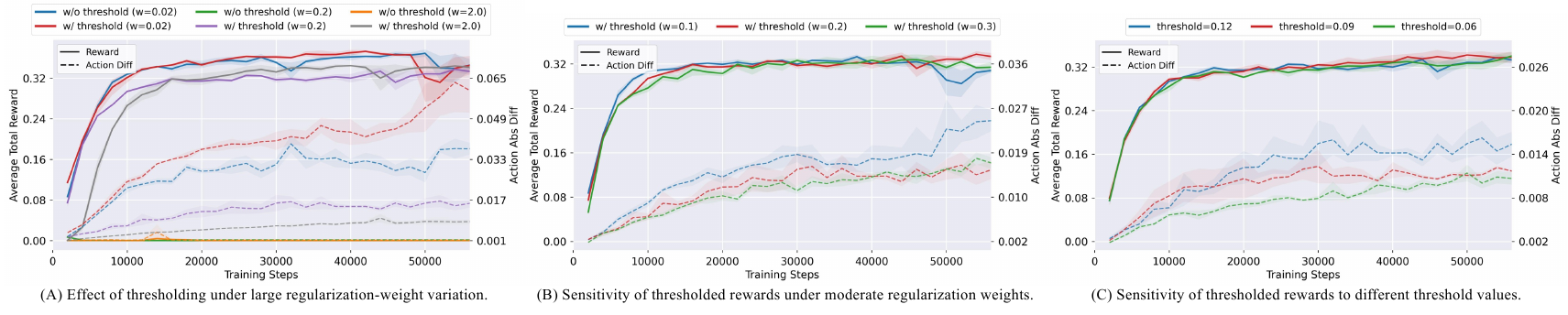}
\caption{Sensitivity analysis of threshold-based rewards. Solid lines denote task-related rewards, and dashed lines denote the mean action difference (action diff).}
  \label{fig:threshold_ablation}
\end{figure}
In this analysis, we evaluate the reward design on the hooking-only task. Because different regularization weights and threshold values change the scale of penalty terms, we report only the task-related reward components. We also report the mean action difference, defined as the average absolute difference between consecutive policy actions, to measure action smoothness. The task reward follows Table~\ref{tab:reward-params}, with weights of 0.3 for both the fine and coarse hooking rewards.

We first vary the action smoothness penalty weight over a wide range, including 0.02, 0.2, and 2.0. As shown in Fig.~\ref{fig:threshold_ablation}A, the threshold-based formulation completes the task under all three weights, even when the penalty weight is 2.0, which is substantially larger than the total hooking reward weight of 0.6. In all cases, the mean action difference also converges, indicating that the policy can improve smoothness without sacrificing task completion. In contrast, without the threshold, the mean action difference remains very small, but the policy succeeds only when the penalty weight is 0.02. For larger weights, the smoothness penalty dominates the task reward and prevents effective learning, and even the 0.02 setting shows instability in the later stage of training.

We further test the threshold-based formulation with action smoothness penalty weights around the default setting, using 0.1, 0.2, and 0.3. As shown in Fig.~\ref{fig:threshold_ablation}B, all three settings achieve comparable task reward and converge to similar ranges of mean action difference. This indicates that the threshold-based reward is not sensitive to moderate changes in the smoothness penalty weight, while still maintaining both task completion and action smoothness.

Finally, we analyze the effect of the threshold value itself. Fig.~\ref{fig:threshold_ablation}C shows that different thresholds have only minor effects on the task reward, while smaller thresholds lead to lower converged mean action difference because the smoothness penalty becomes active earlier.
\end{document}